%% file: main.tex
\pdfoutput=1
\documentclass{article}
\usepackage{iclr2027_conference,times}

\input{math_commands.tex}

\input{useful_package.tex}

\title{$TF^2$-Steering: Action Refinement with Tactile Forecasts and Feedback via Velocity-Field Steering}

\author{%
\textbf{Yiming Chen$^{1,2,3}$ \qquad Xiaofei Wei$^{2,4}$ \qquad Chun Gu$^{2,4}$} \\
\textbf{Wei Xiao$^{2,4}$ \qquad Peidong Liu$^{3}$ \qquad Li Zhang$^{2,4}$\thanks{Corresponding author.}} \\[5pt]
$^1$Zhejiang University \quad $^2$Shanghai Innovation Institute \\
$^3$Westlake University \quad $^4$Fudan University
}
\iclrfinalcopy

\begin{document}

\maketitle
\fancyhead{}

\begin{figure}[!ht]
  \centering
  \includegraphics[width=\textwidth]{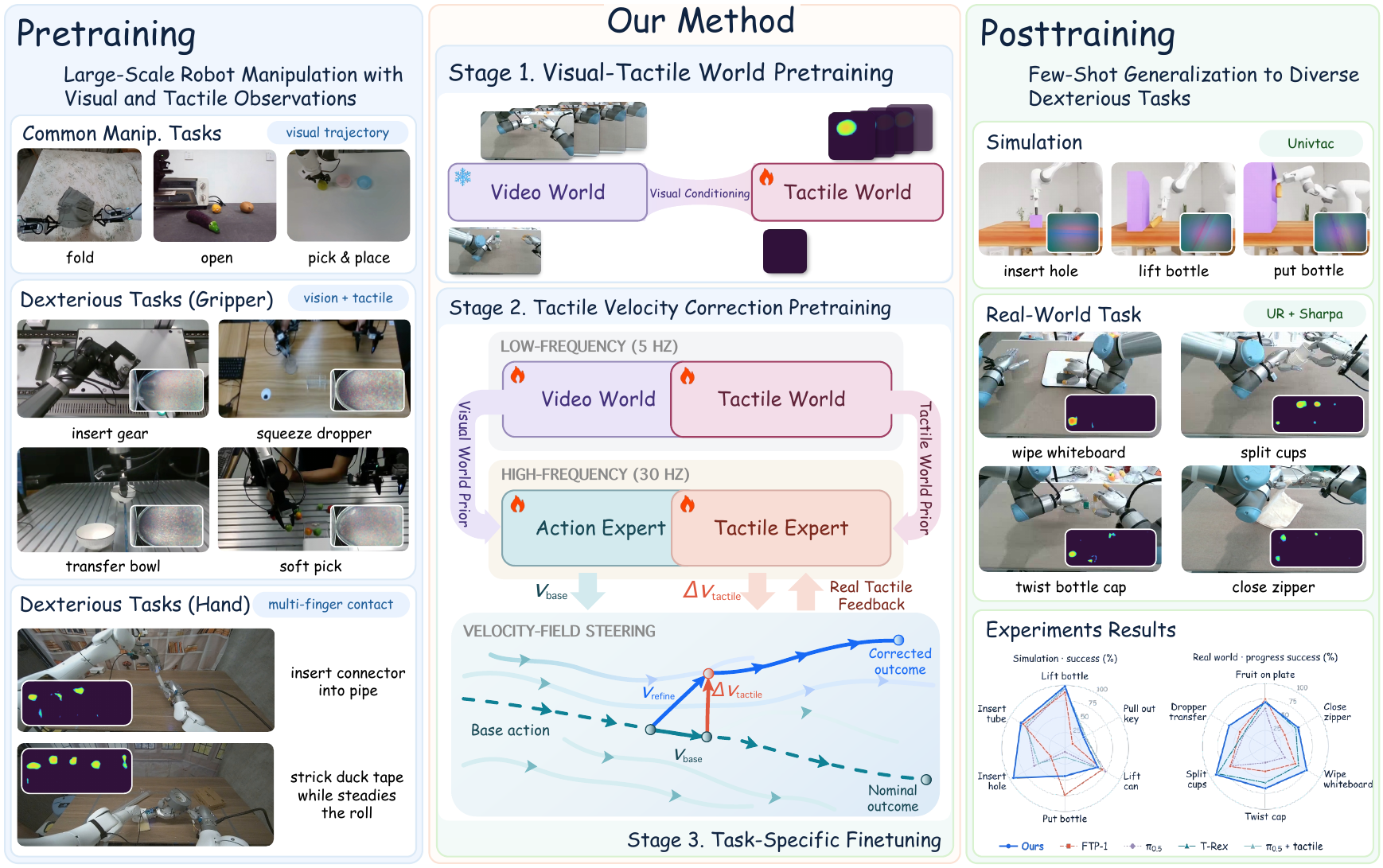}
  \caption{Overview of \methodabbr{}. \textbf{Left:} visual-only and visual--tactile pretraining data. \textbf{Middle:} the tactile expert uses cached forecasts from the tactile world model and real-time tactile feedback to predict delta velocity, refining the action expert's flow-matching velocity. The video world model and action expert remain independently usable without tactile input. \textbf{Right:} task-specific post-training and evaluation in simulation and the real world.}
  \label{fig:teaser}
\end{figure}

\input{sec/0_abstract.tex}
\input{sec/1_introduction.tex}
\input{sec/2_related.tex}
\input{sec/3_method.tex}
\input{sec/4_experiments.tex}
\input{sec/5_conclusions.tex}
\input{sec/6_statements.tex}

\bibliography{main}
\bibliographystyle{iclr2027_conference}

\appendix
\input{sec/appendix.tex}

\end{document}

%% file: math_commands.tex
\usepackage{amsmath,amsfonts,bm}

\def\eqref#1{equation~\ref{#1}}

\def\1{\bm{1}}

\DeclareMathAlphabet{\mathsfit}{\encodingdefault}{\sfdefault}{m}{sl}
\SetMathAlphabet{\mathsfit}{bold}{\encodingdefault}{\sfdefault}{bx}{n}



%% file: useful_package.tex
\newcommand{\methodabbr}{$TF^2$-Steering}

\usepackage{graphicx}
\usepackage{booktabs}
\usepackage{amsmath}
\usepackage{amssymb}
\usepackage{microtype}
\usepackage{xcolor}
\usepackage{colortbl}
\usepackage{url}
\usepackage{hyperref}

\definecolor{ablationblue}{RGB}{232,243,249}

%% file: sec/0_abstract.tex
\begin{abstract}

Tactile sensing plays a crucial role in everyday human interaction. Consider inserting a plug into a tight socket, vision alone provides limited guidance during the final insertion stage, and we instead rely on both anticipated contact and realized tactile feedback to ajust the insertion motion. The same capability is equally important for robots, particularly in dexterous manipulation tasks.
Yet many contemporary robot policies remain vision-dominant. This visual bias, compounded by limited tactile data and sparse contact events, can cause directly incorporated tactile signals to be ignored or even compromise the policy's pretrained capabilities, yielding degrading performance as observed in our experiments.
We present \methodabbr{}, which refines actions via learned residual steering of the flow-matching action expert's velocity field, using a tactile mixture-of-experts architecture to integrate both tactile forecasts and real-time tactile feedback. Together with specific designed three stages training recipe, \methodabbr{} enables tactile signals to correct and, when necessary, redirect the flow-matching action-generation process, while building on the strong action-generation capacity of the vision-based foundation policy.
Extensive experiments across six simulated and six real-world manipulation tasks demonstrate the effectiveness of \methodabbr{}. It outperforms the strongest evaluated baselines by 10.7 percentage points in average simulation success rate and 9.1 percentage points in average real-world progress success rate.
Project page: \url{https://yiming-cc.github.io/tf2-steering/}.

\end{abstract}

%% file: sec/1_introduction.tex
\section{Introduction}
\label{sec:introduction}

Recent robot policy models, particularly vision-language-action (VLA) and world-action (WA) foundation policies~\citep{intelligence2025pi05,ma2026internvla,cai2026xiaomi,zhang2026hy}, have advanced rapidly and demonstrated strong generalization across tasks, objects, and environments. However, their capabilities remain limited in contact-rich manipulation, where task success depends on transient physical interactions that are difficult to infer reliably from visual observations alone. Tactile sensing provides a complementary source of information about these interactions, making it especially valuable for dexterous manipulation. Accordingly, recent studies have begun incorporating tactile signals into robot foundation policies and have reported promising results on dexterous and precision contact-rich tasks~\citep{yu2026forcevla,lou2026dream,li2026forcevla2,fu2026decal,cheng2026omnivtla,yuan2026ftp}. These results highlight tactile sensing as an important direction for extending the capabilities of general-purpose robot policies.

Yet two important questions have not been sufficiently discussed. First, \textbf{how can tactile information be incorporated into a vision-dominant foundation policy?} Robot foundation models are trained on large-scale demonstration data and acquire strong visual action-planning abilities. The temporal sparsity of tactile signals and the limited scale of tactile data relative to visual data may cause a vision-dominant policy to ignore tactile cues when they are naively appended during fine-tuning, or disturb the pretrained visual representation and reduce generalization to settings without tactile sensors~\citep{liu2025factr,chen2026implicitrdp,zhang2026craft}. Recent approaches therefore isolate tactile influence through residual pathways or gating mechanisms~\citep{zhou2026touchworld,li2026vla}. However, Residual pathways typically introduce a separately trained correction branch that can only modify an already generated action, limiting its ability to redirect a suboptimal plan. Gating mechanisms preserve the backbone, but their learned gates may attenuate tactile features before they reach the action head, making the tactile contribution weak or inconsistent. Second, \textbf{should tactile sensing be treated solely as feedback, or also modeled predictively?} Human manipulation relies on both: prior experience allows us to anticipate an appropriate grasp configuration and force from visual context, while realized contact feedback enables online adjustment during execution. Existing studies have explored reactive tactile feedback~\citep{he2025foar,xue2025reactive,niu2026t} and future tactile prediction~\citep{heng2025vitacformer,yuan2026vtam,lou2026dream}, but these roles have not yet been consistently integrated into a single causal action-correction process. We posit that reliable dexterous manipulation depends on both the expected and realized contact for action refinement.

In this work, we introduce \methodabbr{}, a framework that further discusses these two questions. Inspired by recent studies that modify flow-matching velocity fields for controlled generation and policy adaptation~\citep{zhao2024flowturbo,park2024constant,zhang2025flowpolicy,patel2025flowchef}, we incorporate tactile information by steering the velocity field of the flow-matching action expert. 
Specifically, \methodabbr{} augments the original action expert with a tactile expert that attends unidirectionally to its features. Using tactile forecasts and real-time tactile feedback, the tactile expert predicts a delta velocity to refine the base flow-matching velocity during action generation, while the original action expert remains independent of tactile input.
To complement real-time feedback with tactile anticipation, we introduce a tactile world model that predicts future tactile states from the available world context. \methodabbr{} adopts a dual-rate design: the video and tactile world models provide low-frequency planning priors, while the action and tactile experts refine actions at a higher frequency during execution. Tactile forecasts are generated before each action block and held fixed throughout its execution, allowing the tactile expert to steer action generation using incoming real-time feedback alongside these cached forecasts.

Our main contributions are threefold:
\begin{itemize}
    \item We develop a dual-rate framework that combines low-frequency tactile forecasts, cached before execution, with high-frequency tactile feedback for online action refinement. A three-stage training strategy supports both tactile forecasting and the effective use of predicted and real-time tactile information.

    \item We introduce a tactile-guided velocity-field steering algorithm that learns delta-velocity corrections within the flow-matching action-generation process. This design enables local action refinement while retaining the original action expert as an independently usable source of base actions.

    \item We evaluate our framework on six simulated and six real-world manipulation tasks, outperforming the strongest evaluated baselines by 10.7 percentage points in average simulation success rate and 9.1 percentage points in average real-world progress success rate. Ablation studies further support the contributions of tactile forecasting, real-time feedback, and velocity-field steering.
\end{itemize}

%% file: sec/2_related.tex
\section{Related Work}
\label{sec:related}

\noindent\textbf{Velocity-Field Adaptation in Flow Matching.}
Flow Matching models generate samples by integrating a learned time-dependent velocity field. Adapting this field provides a modular way to build upon a pretrained flow model while introducing task-specific control. Existing studies have explored velocity-field adaptation for several purposes. FlowTurbo~\citep{zhao2024flowturbo} and Constant Acceleration Flow~\citep{park2024constant} introduce velocity residuals or acceleration terms to improve trajectory expressiveness and sampling efficiency, while FlowPolicy~\citep{zhang2025flowpolicy} and SCoT~\citep{fan2026scot} enforce velocity consistency for fast generation. Other works, including FlowChef~\citep{patel2025flowchef}, FlowEdit~\citep{kulikov2025flowedit}, and DNAEdit~\citep{xie2026dnaedit}, steer or refine velocity fields to incorporate measurements or perform controlled image editing. Graph Flow Matching~\citep{siddiqui2026graph} and Energy Matching~\citep{balcerak2026energy} further augment flow dynamics with neighborhood or energy-based priors. Our work extends this paradigm to tactile-conditioned action generation by learning an action-space correction field from contact observations.

\noindent\textbf{Visual-tactile Manipulation.}
Tactile sensing complements vision by providing localized cues about contact geometry, force, friction, and slip, which are particularly valuable for dexterous manipulation when visual observations are severely occluded. Early studies incorporated tactile or force feedback into task-specific visuomotor policies, including FoAR~\citep{he2025foar} and Reactive Diffusion Policy~\citep{xue2025reactive}.
More recent work has integrated tactile sensing into general-purpose VLA and WA models. One line of research focuses on tactile-language grounding and semantic alignment, linking tactile observations to task-relevant physical concepts~\citep{hao2025tla,zhang2026vtla,huang2025tactile,cheng2026omnivtla}. Another investigates force-aware action generation and tactile-reactive policy architectures through force-conditioned expert routing~\citep{yu2026forcevla}, hybrid force-position control~\citep{li2026forcevla2}, and slow-fast inference~\citep{niu2026t}. Large-scale tactile policy pretraining has also been explored to facilitate the transfer of manipulation skills across heterogeneous sensors and embodiments~\citep{yuan2026ftp}.
However, in vision-dominant policy models, naively appending tactile inputs can leave tactile cues underutilized and degrade the pretrained policy's ability to plan and generate actions, consistent with prior findings on force underutilization and modality imbalance~\citep{liu2025factr,chen2026implicitrdp,zhang2026craft}. Existing methods address these challenges through contact-aware gating~\citep{zhang2026tacvla}, adaptive tactile injection~\citep{li2026vla}, asymmetric attention~\citep{wu2026tactile,tian2026vt}, and bounded residual action modulation~\citep{wang2026vitar}. Our method applies tactile-driven corrections directly to the flow-matching velocity field, preserving the pretrained policy's visual planning while enabling precise contact-level adjustments.

A complementary line of work predicts future tactile states to learn internal contact grounding, treating tactile prediction either as an auxiliary objective or as a component of a visuotactile world model~\citep{heng2025vitacformer,huang2026taf,yuan2026vtam,zheng2026omnivta,lou2026dream}. TacForcing and TacPAC~\citep{zhou2026tacforcing,ma2026tacpac} further mitigate the temporal mismatch between chunked action generation and tactile feedback available during execution. Our method brings these two functions together through a tactile expert that jointly uses tactile forecasts and real-time tactile feedback to refine the action expert's flow-matching velocity during action generation.

%% file: sec/3_method.tex
\section{Method}
\label{sec:method}

Given a task instruction $\mathcal{I}$, the current multi-view visual observation $\mathbf{o}^{\mathrm{v}}_{t}$, and the current tactile observation $\mathbf{o}^{\mathrm{\tau}}_{t}$, \methodabbr{} produces an action chunk $\mathbf{A}_{t:t+H-1}$, where $t$ denotes the current control step, $H$ is the
planning horizon, and the superscripts $\mathrm{v}$ and $\mathrm{\tau}$ indicate the visual and tactile modalities, respectively.
During closed-loop execution, real-time tactile feedback is passed to the tactile expert to refine the flow-matching velocity for the next unexecuted action block.
In the remainder of this section, we describe the \methodabbr{} model architecture, the asynchronous velocity-field steering algorithm, and the three-stage training procedure.

\begin{figure}[!ht]
  \centering
  \includegraphics[width=\textwidth]{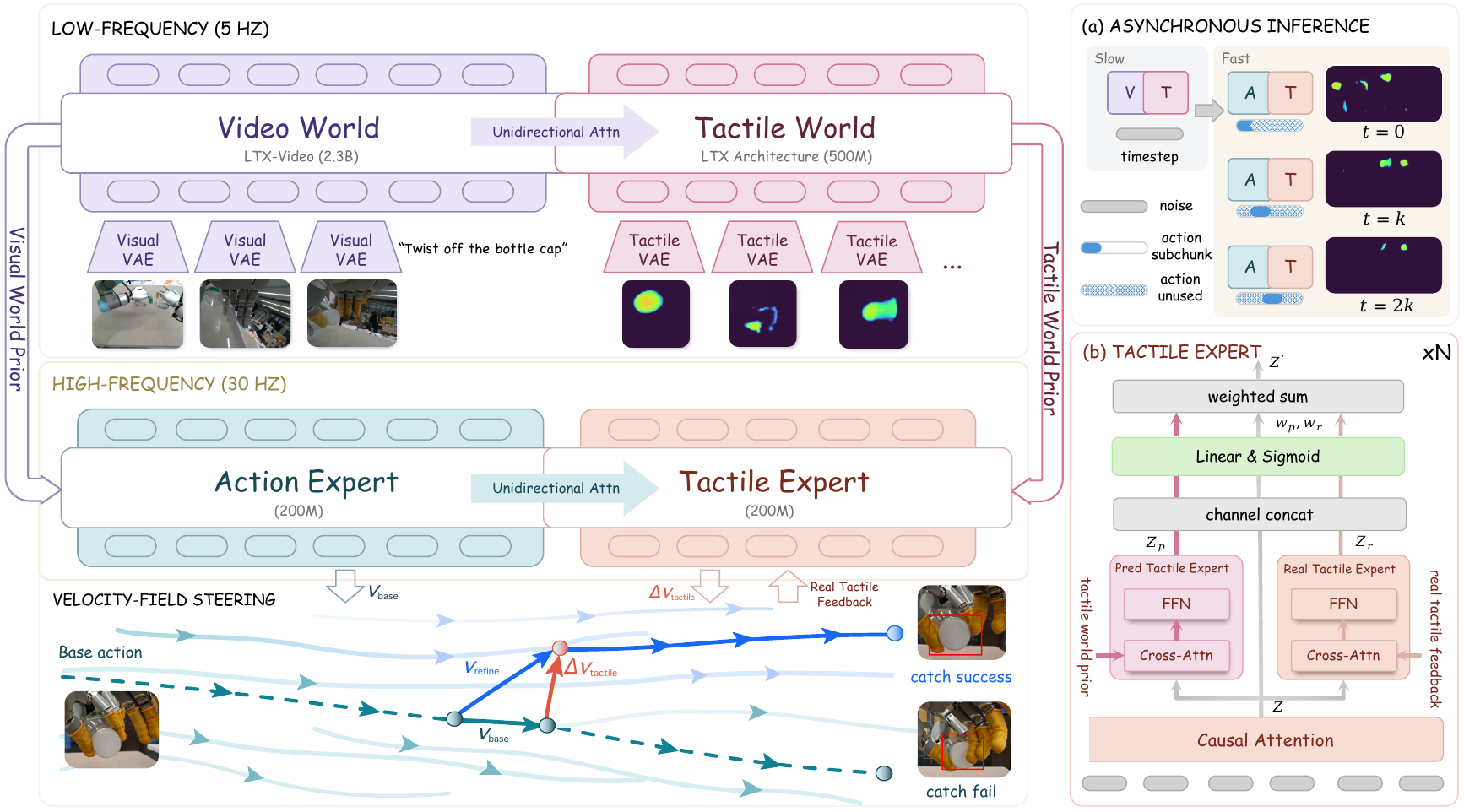}
  \caption{\textbf{Overview of \methodabbr{}.}
  The video and tactile world models provide low-frequency planning priors, while the action and tactile experts perform high-frequency action refinement through learned delta-velocity corrections. Unidirectional attention keeps the original world-action backbone independent of tactile input.
  \textbf{(a)} Asynchronous inference uses incoming tactile feedback to refine subsequent action blocks, with pre-execution tactile forecasts held fixed throughout each block.
  \textbf{(b)} The tactile expert integrates predicted and real-time tactile features through separate subexperts with sigmoid weighting to steer the action expert's flow-matching velocity field.}
  \label{fig:pipeline}
\end{figure}

\subsection{Model Architecture}
As shown in Fig.~\ref{fig:pipeline}, \methodabbr{} consists of a pretrained world-action model, a tactile world model, and a forecast--feedback tactile correction expert. The world-action model provides the visual context and base action generation, while the tactile world model and tactile correction expert enable tactile-guided action refinement.

\noindent\textbf{World-Action Model Backbone.}
The world-action model comprises LTX-Video~\citep{hacohen2024ltx}, a modality compression Q-Former, and a flow-matching action expert. To better accommodate the robot's multi-view observations, we augment LTX-Video with a multi-view cross-attention module. We train the video model with a single-step denoising objective by setting the flow-matching noise ratio to $1$ during training. This design allows the video model to perform inference with a single forward pass. The Q-Former subsequently compresses the visual context into compact query features, which condition the action expert for action generation. As a result, the model generates an action chunk in $240\,\mathrm{ms}$ on a single NVIDIA RTX 4090. Additional training and implementation details are provided in
Appendix~\ref{app:wa-pretraining}.

\noindent\textbf{Tactile World Model.}
The tactile world model is built on the LTX-Video architecture, but uses a smaller model with approximately $500$M parameters. We remove the multi-view cross-attention module, which is unnecessary for tactile images, and reduce the model width and depth while retaining roughly one-third of the original layers. The tactile world model is connected to LTX-Video through a unidirectional attention pathway: the tactile predictor can attend to visual features, whereas the visual stream cannot attend to tactile features. During training, the model receives the current tactile image and predicts a future tactile observation using the same single-step denoising formulation as the video model.
To provide a suitable latent representation for tactile inputs, we first fine-tune the LTX-Video VAE on tactile videos and then use the resulting VAE to encode tactile images. At inference time, rather than decoding the predicted tactile image, we use the feature from the final layer of the tactile world model as the tactile-world prior for the downstream tactile expert. This feature-level interface reduces inference cost while retaining richer task-related context than a decoded tactile image.

\noindent\textbf{Forecast-Feedback Tactile Correction Expert.}
We introduce a forecast--feedback tactile correction expert, referred to as the \emph{tactile expert} hereafter, to integrate both predicted tactile features from the tactile world model and real-time tactile features. The tactile expert attends unidirectionally to the action expert's features, with stop-gradient applied before these features enter the tactile expert. This allows the tactile expert to access action context while keeping tactile inputs outside the action expert.
As detailed in Fig.~\ref{fig:pipeline}(b), each tactile expert block contains separate predicted-tactile and real-tactile subexperts, inspired by mixture-of-experts architectures~\citep{fedus2022switch,jiang2024mixtral}. We use sigmoid weighting to modulate the two branches independently, without constraining their weights to sum to one.

The tactile expert and the action expert receive the same noised action $\widetilde{\mathbf{a}}_{s}$ and flow-matching timestep $s$. Let $\mathbf{v}^{\mathrm{base}}_{\theta}$ denote the base flow-matching velocity produced by the action expert, $\Delta\mathbf{v}_{\phi}$ denote the correction predicted by the tactile expert, and $\mathbf{v}^{\star}$ denote the flow matching target. As illustrated in the lower-left portion of Fig.~\ref{fig:pipeline}, the tactile expert steers action generation by refining the base velocity:
\begin{equation}
\mathbf{v}^{\mathrm{corr}}
=
\mathbf{v}^{\mathrm{base}}_{\theta}
+
\Delta\mathbf{v}_{\phi}.
\end{equation}
The action and tactile experts are trained with separately computed flow
matching losses:
\begin{equation}
\mathcal{L}_{\mathrm{act}}
=
\left\|
\mathbf{v}^{\mathrm{base}}_{\theta}
-
\mathbf{v}^{\star}
\right\|_{2}^{2},
\qquad
\mathcal{L}_{\mathrm{steer}}
=
\left\|
\mathbf{v}^{\star}
-
\left(
\operatorname{sg}\!\left[
\mathbf{v}^{\mathrm{base}}_{\theta}
\right]
+
\Delta\mathbf{v}_{\phi}
\right)
\right\|_{2}^{2},
\end{equation}
where $\operatorname{sg}[\cdot]$ denotes the stop-gradient operator. Together with the stop-gradient on the base velocity in the steering objective, detaching the action features prevents the steering loss from updating the action expert, which remains trainable through its own action loss.

\subsection{Asynchronous Velocity-Field Steering}
To support high-frequency tactile feedback and action correction, \methodabbr{} uses a two-rate control system. The video world model and tactile world model form a low-frequency planning system that generates long-horizon visual and tactile predictions, whereas the action expert and tactile expert form a high-frequency steering system that updates the action sequence during execution. 

Figure~\ref{fig:pipeline}(a) illustrates the asynchronous generation and execution of an action chunk containing $H$ control steps, executed in blocks of $k$ steps. The low-frequency system performs inference once to generate the chunk, and the resulting visual and tactile world priors remain fixed throughout its execution and all subsequent high-frequency updates. After each subchunk $k$ is executed, the high-frequency system receives the latest tactile feedback and regenerates the remaining actions. During every flow-matching iteration of this regeneration process, the tactile expert continues to correct the action expert's base velocity. Actions that have already been executed are discarded, and the robot executes only the newly generated actions that remain unexecuted.

During training, we randomly sample a delay $0 \leq d < H$ to simulate the delayed arrival of tactile feedback in the high-frequency system. At time $t$, the video world model receives $\mathbf{o}^{\mathrm{v}}_{t}$ and the tactile world model receives $\mathbf{o}^{\mathrm{\tau}}_{t}$, while the real-time tactile observation $\mathbf{o}^{\mathrm{\tau,real}}_{t+d}$ is provided to the tactile expert after the sampled delay. The tactile correction loss is evaluated only within a steering window after the delay. Specifically, for
\begin{equation}
\mathcal{W}_{d}
=
\left\{
i \,\middle|\, d \leq i < \min(d+W,H)
\right\},
\end{equation}
we define
\begin{equation}
\mathcal{L}_{\mathrm{steer}}
=
\frac{1}{|\mathcal{W}_{d}|}
\sum_{i\in\mathcal{W}_{d}}
\left\|
\mathbf{v}^{\star}_{i}
-
\left(
\operatorname{sg}\!\left[
\mathbf{v}^{\mathrm{base}}_{\theta,i}
\right]
+
\Delta\mathbf{v}_{\phi,i}
\right)
\right\|_{2}^{2}.
\end{equation}

We provide the tactile expert with a relative-position encoding $\rho_i=i-d$ to locate each action relative to the feedback time. Together with the windowed loss, this encoding associates tactile feedback with the corresponding action window. The block length $k$ can be adjusted at inference time to control the refinement frequency.

\subsection{Training}
\methodabbr{} uses large-scale tactile pretraining for both the tactile world
model and the tactile expert, providing a stronger initialization for
generalization across diverse contact conditions.

\noindent\textbf{Tactile Prediction Pretraining.}
In the first stage, we pretrain the tactile world model on large-scale tactile
datasets, including OpenNeoData~\citep{team2026n_0vtla}, the T-REX
dataset~\citep{niu2026t}, and FTP-1-Sharpa~\citep{yuan2026ftp}. The video world
model provides the visual conditioning feature
$\mathbf{h}^{\mathrm{v}}_{t}$ but is kept fixed throughout this stage. Given the
current tactile observation $\mathbf{o}^{\mathrm{\tau}}_{t}$, a noised VAE latent
$\widetilde{\mathbf{x}}^{\mathrm{\tau}}_{s}$ of a future tactile observation, and the
flow-matching timestep $s$, the tactile world model is trained with
\begin{equation}
\mathcal{L}_{\mathrm{pred}}^{\mathrm{\tau}}
=
\mathbb{E}
\left[
\left\|
\mathbf{v}^{\mathrm{\tau},\star}
-
\mathbf{v}^{\mathrm{\tau}}_{\psi}
\left(
\widetilde{\mathbf{x}}^{\mathrm{\tau}}_{s},
s
\mid
\mathbf{o}^{\mathrm{\tau}}_{t},
\mathbf{h}^{\mathrm{v}}_{t}
\right)
\right\|_{2}^{2}
\right],
\end{equation}
where $\mathbf{v}^{\mathrm{\tau},\star}$ denotes the flow-matching target in
the VAE latent space of the future tactile observation.

\noindent\textbf{Tactile Velocity Steering Pretraining.}
In the second stage, we train all modules end-to-end on the same tactile data to
enable the tactile expert to use both predicted tactile features and real-time
tactile feedback for action correction. To preserve the generalization of the
world-action backbone, we additionally mix in $10\%$ tactile-free samples from
the original world-action pretraining data. Let
$\mathcal{L}_{\mathrm{pred}}^{\mathrm{v}}$ denote the video-world prediction
loss. The joint training objective is
\begin{equation}
\mathcal{L}_{\mathrm{joint}}
=
\mathcal{L}_{\mathrm{pred}}^{\mathrm{v}}
+
\mathcal{L}_{\mathrm{pred}}^{\mathrm{\tau}}
+
\mathcal{L}_{\mathrm{act}}
+
\lambda_{\mathrm{steer}}
\mathcal{L}_{\mathrm{steer}}.
\end{equation}
Here, $\lambda_{\mathrm{steer}}$ controls the contribution of the steering loss.
Further details of the data mixture and training configuration are provided in
Appendix~\ref{app:tactile-pretraining}.

\noindent\textbf{Task-Specific Fine-Tuning.}
In the final stage, we retain the same training objective as in the tactile
velocity steering pretraining stage and perform task-specific post-training on
both simulated and real-robot data. This procedure adapts the pretrained model
to the target manipulation tasks and yields clear performance gains over
training without tactile pretraining.

%% file: sec/4_experiments.tex
\section{Experiments}
\label{sec:experiments}

\begin{figure}[!ht]
    \centering
    \includegraphics[width=\textwidth]{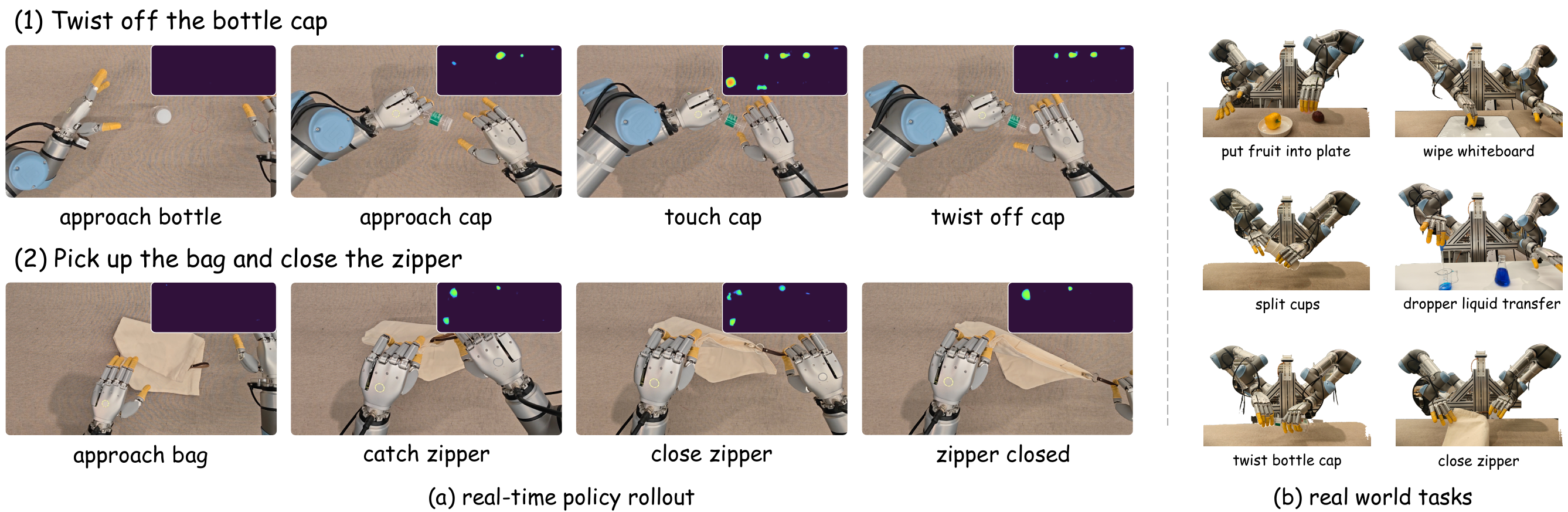}
    \caption{Real-world manipulation tasks and example execution sequences.
    \textbf{(a)} Bottle-cap twisting and zipper closing, with tactile
    visualizations shown as insets.
    \textbf{(b)} Task scenes for fruit placement, whiteboard wiping, cup
    separation, liquid transfer with a dropper, bottle-cap twisting, and
    zipper closing.}
    \label{fig:real-world-tasks}
\end{figure}

\begin{table}[t]
    \centering
    \hypersetup{hidelinks}
    \caption{\textbf{Simulation results on UniVTAC.} Success rate
    (\%, $\uparrow$) on six tasks; Avg. denotes the macro-average.
    \textbf{Bold} and \underline{underlined} values indicate the best and
    second-best results in each column, respectively, including ties.
    $^{*}$ACT and VITaL results are taken from UniVTAC~\citep{chen2026univtac}.}
    \label{tab:univtac-main}
    \small
    \setlength{\tabcolsep}{2.8pt}
    \renewcommand{\arraystretch}{1.15}
    \begin{tabular*}{\textwidth}{@{\extracolsep{\fill}}lccccccc@{}}
        \toprule
        Method
        & \shortstack{Lift\\Bottle}
        & \shortstack{Pull Out\\Key}
        & \shortstack{Lift\\Can}
        & \shortstack{Put\\Bottle}
        & \shortstack{Insert\\Hole}
        & \shortstack{Insert\\Tube}
        & Avg. \\
        \midrule
        ACT$^{*}$~\citep{zhao2023learning}
            & 71.0 & \underline{46.0} & 29.0 & 31.0 & 24.0 & 56.0 & 42.8 \\
        VITaL$^{*}$~\citep{george2025vital}
            & 72.0 & \textbf{47.0} & 8.0 & 32.0 & 25.0 & 34.0 & 36.3 \\
        FTP-1~\citep{yuan2026ftp}
            & 88.0 & 14.0 & \underline{68.0} & \textbf{74.0} & 26.0 & \underline{76.0} & \underline{57.7} \\
        $\pi_{0.5}$~\citep{intelligence2025pi05}
            & 94.0 & 36.0 & \textbf{74.0} & 6.0 & \underline{56.0} & 68.0 & 55.7 \\
        $\pi_{0.5}$ + tactile
            & \underline{96.0} & 38.0 & \underline{68.0} & 14.0 & 48.0 & 72.0 & 56.0 \\
        \midrule
        \methodabbr{} (ours)
            & \textbf{98.0} & 34.0 & 60.0 & \underline{44.0} & \textbf{94.0} & \textbf{80.0} & \textbf{68.3} \\
        \bottomrule
    \end{tabular*}
\end{table}

\begin{table}[t]
    \centering
    \hypersetup{hidelinks}
    \caption{\textbf{Real-world results.} Progress success rate
    (\%, $\uparrow$) over 20 trials per task; Avg. denotes the macro-average
    over the six tasks. \textbf{Bold} and \underline{underlined} values
    indicate the best and second-best results in each column, respectively.}
    \label{tab:real-robot-main}
    \small
    \setlength{\tabcolsep}{2.8pt}
    \renewcommand{\arraystretch}{1.15}
    \begin{tabular*}{\textwidth}{@{\extracolsep{\fill}}lccccccc@{}}
        \toprule
        Method
        & \shortstack{Fruit\\on Plate}
        & \shortstack{Close\\Zipper}
        & \shortstack{Wipe\\Whiteboard}
        & \shortstack{Twist\\Cap}
        & \shortstack{Split\\Cups}
        & \shortstack{Dropper\\Transfer}
        & Avg. \\
        \midrule
        ViTacFormer~\citep{heng2025vitacformer}
            & 12.5 & 4.0 & 2.0 & 9.0 & 7.0 & 2.0 & 6.1 \\
        $\pi_{0.5}$~\citep{intelligence2025pi05}
            & 60.0 & 18.0 & 36.0 & 26.0 & 62.0 & 39.0 & 40.2 \\
        $\pi_{0.5}$ + tactile
            & 58.0 & 12.0 & 39.0 & 22.0 & 57.0 & 23.0 & 35.2 \\
        T-Rex~\citep{niu2026t}
            & \underline{72.5} & \underline{56.0} & \underline{61.0} & \underline{58.0} & \underline{87.0} & 40.0 & \underline{62.4} \\
        FTP-1~\citep{yuan2026ftp}
            & \textbf{75.0} & 36.0 & 55.0 & 40.0 & 64.0 & \underline{46.0} & 52.7 \\
        \midrule
        \methodabbr{} (ours)
            & 70.0 & \textbf{60.0} & \textbf{76.0} & \textbf{67.0} & \textbf{90.0} & \textbf{66.0} & \textbf{71.5} \\
        \bottomrule
    \end{tabular*}
\end{table}

\begin{table}[t]
    \centering
    \hypersetup{hidelinks}
    \caption{\textbf{Model design ablations.} Real-world progress success rate
    (\%, $\uparrow$) on three manipulation tasks; Avg. denotes the macro-average.
    $\Delta$ Avg. reports the change relative to the full model in percentage
    points (pp), computed before rounding; negative values indicate a decline.
    \textbf{Bold} indicates the best result in each task and average column,
    including ties.}
    \label{tab:ablation}
    \small
    \setlength{\tabcolsep}{2.8pt}
    \renewcommand{\arraystretch}{1.15}
    \begin{tabular*}{\textwidth}{@{\extracolsep{\fill}}lccccc@{}}
        \toprule
        Configuration
        & \shortstack{Twist\\Cap}
        & \shortstack{Close\\Zipper}
        & \shortstack{Split\\Cups}
        & Avg.
        & \shortstack{$\Delta$ Avg.\\(pp)} \\
        \midrule
        \methodabbr{} (full model)
            & \textbf{67.0} & \textbf{60.0} & \textbf{90.0} & \textbf{72.3} & -- \\
        \midrule
        \rowcolor{black!5}
        \multicolumn{6}{@{\hspace{4pt}}l}{\textit{Tactile modalities}} \\
        \quad w/o predicted tactile
            & 61.0 & 58.0 & \textbf{90.0} & 69.7 & $-2.7$ \\
        \quad w/o real-time tactile
            & 54.0 & 49.0 & 86.0 & 63.0 & $-9.3$ \\
        \quad w/o both tactile modalities
            & 51.0 & 50.0 & 84.0 & 61.7 & $-10.7$ \\
        \addlinespace[2pt]
        \rowcolor{black!5}
        \multicolumn{6}{@{\hspace{4pt}}l}{\textit{Tactile integration strategy}} \\
        \quad Action-space residual correction
            & 53.0 & 49.0 & 82.0 & 61.3 & $-11.0$ \\
        \addlinespace[2pt]
        \rowcolor{black!5}
        \multicolumn{6}{@{\hspace{4pt}}l}{\textit{Architecture}} \\
        \quad w/o asynchronous inference
            & 56.0 & 54.0 & \textbf{90.0} & 66.7 & $-5.7$ \\
        \quad w/o attention masking \& stop-gradient
            & 49.0 & 51.0 & 88.0 & 62.7 & $-9.7$ \\
        \bottomrule
    \end{tabular*}
\end{table}

\subsection{Experiment Setup}

\noindent\textbf{Tasks and Evaluation Metrics.}
We evaluate \methodabbr{} in both simulation and real-world experiments.
For simulation, we use UniVTAC~\citep{chen2026univtac}, a benchmark for contact-rich manipulation that includes tasks such as lifting a bottle and pulling out a key.
Since some UniVTAC tasks support only single-camera observations, we train two checkpoints for each method: one for the single-view task group and one for the multi-view task group (Appendix~\ref{app:simulation-evaluation}).
For real-world experiments, we use a bimanual platform comprising two 6-DoF UR5e arms, each equipped with a 22-DoF Sharpa dexterous hand (Appendix~\ref{app:real-robot-setup}).
We collect 100 demonstrations for each of six tasks: putting fruit on a plate, wiping a whiteboard, splitting cups, transferring liquid with a dropper, twisting a bottle cap, and closing a zipper (Fig.~\ref{fig:real-world-tasks}b).
All baselines undergo task-specific training on the collected demonstrations, with pretrained models adapted through post-training.
We evaluate each method over 20 trials per real-world task and report progress success rate to capture partial task completion and provide a more fine-grained assessment of manipulation performance. The scoring rules are detailed in Appendix~\ref{app:real-world-evaluation}.

\noindent\textbf{Baseline.}
We compare \methodabbr{} with five baselines:
(1) \textbf{ViTacFormer}~\citep{heng2025vitacformer}, an ACT-based policy trained separately for each task that predicts future tactile force and torque signals to support action generation;
(2) \textbf{$\pi_{0.5}$}, a vision-language-action (VLA) model pretrained on large-scale, diverse data;
(3) \textbf{$\pi_{0.5}$ + tactile}, our implementation that concatenates all tactile images into a single composite image and supplies it as an additional image input to $\pi_{0.5}$, testing the effect of directly incorporating tactile observations on task performance;
(4) \textbf{T-Rex}~\citep{niu2026t}, a tactile-reactive VLA system that incorporates high-frequency tactile feedback during inference through cascaded flow matching for action refinement;
and (5) \textbf{FTP-1}~\citep{yuan2026ftp}, a foundation tactile policy pretrained on large-scale tactile data that unifies heterogeneous tactile signals and integrates them into a VLA architecture.
Further details on baseline training and adaptation are provided in Appendix~\ref{app:real-world-training}.

\subsection{Main Results}
\noindent\textbf{Simulation.}
As shown in Table~\ref{tab:univtac-main}, \methodabbr{} achieves an average success rate of 68.3\%, outperforming the strongest baseline, FTP-1, by 10.6 percentage points.
The gains are particularly pronounced on Insert Hole, where our method achieves 94.0\% success compared with 56.0\% for the strongest baseline, while also attaining the highest success rate on Insert Tube.
These results support the effectiveness of tactile-guided action refinement for precise insertion.
In comparison, directly incorporating tactile images into $\pi_{0.5}$ yields little improvement in average success rate (56.0\% versus 55.7\%), suggesting that simply providing tactile observations is insufficient for effective use during action generation.
This may reflect the difficulty of translating tactile signals supplied as image observations into appropriate action corrections, given that the model's large-scale pretraining primarily emphasizes vision-driven action planning.

\noindent\textbf{Real World Tasks.}
Table~\ref{tab:real-robot-main} shows that \methodabbr{} achieves an average progress success rate of 71.5\%, exceeding the strongest baseline, T-Rex, by 9.1 percentage points.
Our method performs particularly well on bottle-cap twisting, liquid transfer with a dropper, and zipper closing.
As illustrated in Fig.~\ref{fig:real-world-tasks}a, these tasks require precise contact interactions under substantial visual occlusion, including establishing a secure grasp before twisting a cap, regulating pressure when squeezing and releasing a dropper, and grasping and pulling a zipper slider.
The relatively low scores of $\pi_{0.5}$ on these tasks highlight the limitations of relying on visual observations alone.
Although FTP-1 improves upon $\pi_{0.5}$, it remains substantially behind our method on bottle-cap twisting (40.0\% versus 67.0\%), consistent with the value of refining actions using real-time tactile feedback.

T-Rex performs competitively across tasks, but its fixed refinement interval in our evaluation, with tactile updates every four action steps, limits flexibility across tasks with different feedback requirements.
In our rollouts, this configuration leads to prolonged execution on tasks such as whiteboard wiping.
\methodabbr{} allows the refinement frequency to be adjusted at inference time, enabling task-dependent trade-offs between tactile responsiveness and execution overhead.

ViTacFormer performs poorly across the evaluated tasks, suggesting that pretrained representations are valuable for learning multistep dexterous manipulation from limited task-specific demonstrations.
Meanwhile, $\pi_{0.5}$ + tactile performs worse than $\pi_{0.5}$ in the real-world evaluation, with average progress success rate decreasing from 40.2\% to 35.2\%.
We hypothesize that adding the composite tactile image introduces a distribution shift: the resulting four-image input differs from the pretraining configuration and may hinder the transfer of pretrained visual representations.

\subsection{Ablation Studies}
We examine the contributions of model design and tactile pretraining to the performance of \methodabbr{}.

\begin{figure}[t]
    \centering
    \includegraphics[width=\textwidth]{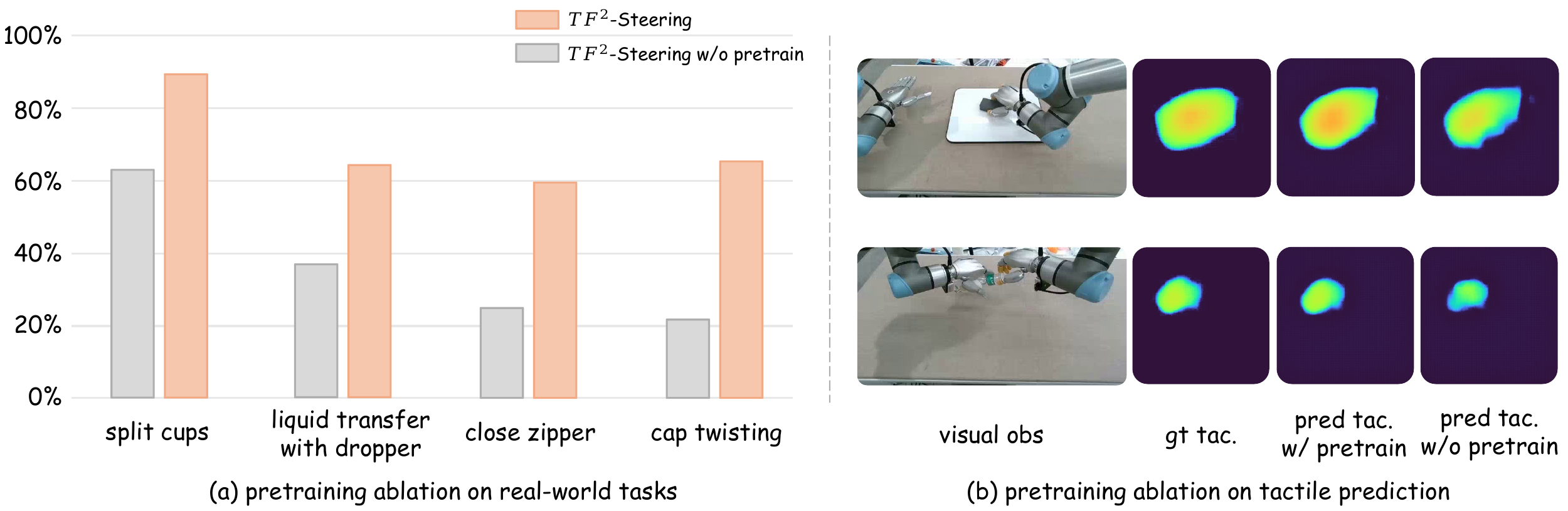}
    \caption{Effect of tactile pretraining.
    \textbf{(a)} Real-world progress success rates with and without tactile
    pretraining on four manipulation tasks.
    \textbf{(b)} Qualitative comparison of tactile forecasts. From left to
    right: visual observation, ground-truth tactile observation, and tactile
    predictions from models with and without tactile pretraining.}
    \label{fig:pretraining-ablation}
\end{figure}

\noindent\textbf{Ablation on Model Design.}
We conduct real-world ablations along three dimensions: tactile modalities, tactile integration strategies, and architectural choices.
As shown in Table~\ref{tab:ablation}, removing both tactile modalities reduces the average progress success rate from 72.3\% to 61.7\%.
Using predicted tactile alone raises this score to 63.0\%, while using real-time tactile alone yields a larger improvement to 69.7\%.
Combining both modalities achieves the highest average performance, suggesting that tactile forecasts and real-time feedback provide complementary information for dexterous manipulation.

For tactile integration, action-space residual correction achieves an average score of 61.3\%, comparable to the variant without tactile inputs (61.7\%).
This result suggests that, under our evaluation settings, correcting an already generated action sequence offers limited benefits compared with refining the flow-matching velocity throughout action generation.

At the architectural level, removing asynchronous inference reduces performance on bottle-cap twisting from 67.0\% to 56.0\%, supporting the value of frequent tactile-guided action updates for precise contact interactions.
We also jointly ablate the attention mask and stop-gradient operation between the action expert and tactile expert.
The attention mask prevents the action expert from accessing tactile features, while stop-gradient prevents the tactile correction loss from updating the action expert.
Together, these constraints preserve the separation between base action generation and tactile-dependent delta-velocity refinement.
Removing both reduces the average score to 62.7\%, close to the variant without tactile inputs.
This degradation may arise from the dominance of the action expert's pretrained visual representations during joint optimization, limiting the contribution of tactile information to action refinement.
Further details of the model-design ablation configurations are provided in Appendix~\ref{app:model-design-ablation}.

\noindent\textbf{Ablation on Pretraining.}
Our three-stage training procedure adapts a pretrained world-action model to tactile-guided manipulation through tactile prediction pretraining, tactile velocity steering pretraining, and task-specific post-training.
Tactile pretraining initializes the tactile world model's ability to forecast future tactile observations and teaches the tactile expert to use predicted and real-time tactile information to produce delta-velocity corrections.
As shown in Fig.~\ref{fig:pretraining-ablation}a, tactile pretraining improves progress success rates across all four evaluated tasks, demonstrating its value beyond task-specific post-training alone.
The pretraining ablation configuration is detailed in Appendix~\ref{app:pretraining-ablation}.
Figure~\ref{fig:pretraining-ablation}b further shows that training the tactile world model only on task-specific post-training data produces poor tactile forecasts.
This qualitative comparison suggests that the limited downstream data are insufficient to learn accurate future tactile prediction, motivating pretraining on a larger and more diverse tactile dataset.

%% file: sec/5_conclusions.tex
\section{Conclusion}
\label{sec:conclusion}

We presented \methodabbr{}, a dual-rate framework that combines cached tactile forecasts and real-time feedback for contact-rich manipulation.
A tactile expert predicts delta-velocity corrections to steer the action expert's flow-matching velocity during action generation.
Simulation and real-world experiments, supported by ablations, demonstrate the effectiveness of this refinement strategy and the value of tactile pretraining.

%% file: sec/6_statements.tex
\subsection*{AI Use Statement}
We used generative AI tools, including OpenAI Codex, to polish the manuscript's
language and assist with code writing during implementation. The authors
reviewed and revised the AI-assisted text and reviewed and tested the
AI-assisted code. The authors take full responsibility for the final
manuscript and implementation.

\subsection*{Ethics Statement}
This work investigates contact-rich robotic manipulation in simulation and
on a laboratory tabletop platform. A central ethical consideration is the
risk of injury or property damage from unintended robot contact. The
reported task performance does not establish safety guarantees for operation
around people or in untested environments. Deployment beyond the evaluated
settings would require additional risk assessment, appropriate hardware and
software safeguards, and human oversight. Reuse or release of demonstration
data should respect the applicable data licenses and protect the privacy of
teleoperators and bystanders.

\subsection*{Reproducibility Statement}
Section~\ref{sec:method} describes the model architecture, velocity-field
steering procedure, and training objectives. Section~\ref{sec:experiments}
specifies the evaluated tasks, baselines, and metrics.
Appendix~\ref{app:pretraining} documents the pretraining data mixtures,
action and state representations, and optimization settings.
Appendix~\ref{app:experimental-details} provides the robot and sensor setup,
demonstration collection procedure, baseline adaptation details, and
evaluation protocols, including the task-specific progress-scoring rules.
Appendix~\ref{app:ablation-details} specifies the ablation configurations.
Representative execution sequences are provided in
Appendix~\ref{app:policy-rollouts}, with full executions in the supplementary
videos.

%% file: sec/appendix.tex
\clearpage
\section{Appendix}
\label{sec:appendix}

\raggedbottom
\setlength{\textfloatsep}{12pt plus 2pt minus 2pt}
\setlength{\intextsep}{12pt plus 2pt minus 2pt}
\setlength{\floatsep}{12pt plus 2pt minus 2pt}

\subsection{Pretraining Details}
\label{app:pretraining}

\subsubsection{World-Action Backbone Pretraining}
\label{app:wa-pretraining}

\noindent\textbf{Data and training setup.}
We pretrain the world-action backbone end-to-end on a mixture of
AgiBot World~\citep{bu2025agibot}, RoboMIND~\citep{hou2025robomind},
Open X-Embodiment~\citep{o2024open}, and RoboCOIN~\citep{wu2025robocoin}.
The video world model, modality compression Q-Former, and action expert
are jointly optimized for visual prediction and action generation.
This stage establishes the backbone before tactile-specific pretraining.
We use the action annotations supplied by each dataset, which may contain
joint-space commands, end-effector (EEF) commands, or both. These annotations
are mapped to a shared 128-dimensional representation described below.

Visual observations are resized and center-cropped to
$192\times256$ pixels. Future video frames are sampled at a target rate of
$15$ frames per second, with $24$ future frames per training sample.
The action expert predicts a sequence of $50$ actions conditioned on visual
features and the robot state. This horizon specifies backbone pretraining;
the execution blocks used for downstream tactile refinement are defined
separately in the main method. Action sequences retain the source datasets'
sampling rates, without a common action-rate resampling target.
Normalization statistics are maintained by dataset identifier.

We train for $100{,}000$ optimizer steps with a global batch size of $512$.
We use AdamW with a base learning rate of $10^{-5}$ and an action-expert
learning rate of $10^{-4}$, followed by cosine decay with a minimum learning
rate of $10^{-6}$ after $1{,}000$ warmup steps. Delay augmentation is applied
with probability $0.5$, with uniformly sampled delays up to $25$ action steps.
Table~\ref{tab:wa-pretraining-config} summarizes the main configuration.

\begin{table}[!htb]
    \centering
    \caption{\textbf{World-action backbone pretraining configuration.}
    The training run uses $100{,}000$ optimizer steps and a global batch size
    of $512$. The action horizon and delay settings refer to backbone
    pretraining.}
    \label{tab:wa-pretraining-config}
    \small
    \setlength{\tabcolsep}{6pt}
    \renewcommand{\arraystretch}{1.0}
    \begin{tabular*}{\textwidth}{@{\extracolsep{\fill}}ll@{}}
        \toprule
        Setting & Value \\
        \midrule
        \multicolumn{2}{@{}l}{\textit{Inputs and architecture}} \\
        Image resolution ($H\times W$) & $192\times256$ \\
        Future video frames / target frame rate & $24$ / $15$ fps \\
        Action horizon & $50$ steps \\
        Action / state dimension & $128$ / $128$ \\
        Video transformer (layers / heads / head dimension) & $28$ / $32$ / $64$ \\
        Multi-view cross-attention & Enabled \\
        Action transformer (layers / hidden dimension / dropout) & $16$ / $1024$ / $0.2$ \\
        \midrule
        \multicolumn{2}{@{}l}{\textit{Optimization}} \\
        Optimizer / $(\beta_1,\beta_2)$ & AdamW / $(0.9,0.95)$ \\
        Training steps / global batch size & $100{,}000$ / $512$ \\
        Base / action-expert learning rate & $10^{-5}$ / $10^{-4}$ \\
        Learning-rate schedule & Cosine decay; minimum $10^{-6}$ \\
        Warmup steps / random seed & $1{,}000$ / $42$ \\
        AdamW $\epsilon$ / weight decay & $10^{-8}$ / $10^{-8}$ \\
        Gradient clipping threshold / accumulation steps & $1.0$ / $1$ \\
        \midrule
        \multicolumn{2}{@{}l}{\textit{Regularization and augmentation}} \\
        State / caption dropout probability & $0.1$ / $0.06$ \\
        Delay augmentation probability & $0.5$ \\
        Maximum delay / sampling distribution & $25$ action steps / uniform \\
        Temporal scaling / mirror augmentation & Disabled / disabled \\
        \bottomrule
    \end{tabular*}
\end{table}

\noindent\textbf{Canonical action and state representation.}
To accommodate different embodiments and annotation formats, we use a fixed
128-dimensional layout for both actions and proprioceptive states.
Table~\ref{tab:canonical-128} gives the slot allocation using zero-based,
inclusive indices. Each group has a fixed capacity; only coordinates
supported by the dataset and embodiment are populated. Joint-space and EEF
annotations occupy separate groups, so datasets providing both can retain
both representations in the same vector. TCP groups are populated when
the dataset supplies EEF annotations and are otherwise zero-filled and
masked out. State and action validity masks
are constructed separately according to their available annotations.

\begin{table}[!htb]
    \centering
    \caption{\textbf{Canonical 128-dimensional action and state layout.}
    Indices are zero-based and inclusive. Group dimensions indicate slot
    capacity, rather than the number of active degrees of freedom for every
    embodiment. TCP denotes the tool center point.}
    \label{tab:canonical-128}
    \small
    \setlength{\tabcolsep}{6pt}
    \renewcommand{\arraystretch}{1.15}
    \begin{tabular*}{\textwidth}{@{\extracolsep{\fill}}lcl@{}}
        \toprule
        Indices & Dimension & Assigned quantity \\
        \midrule
        $0$--$9$ & $10$ & Left-arm joint representation \\
        $10$--$19$ & $10$ & Left TCP / end-effector representation \\
        $20$ & $1$ & Left gripper \\
        $21$--$49$ & $29$ & Left dexterous hand \\
        $50$--$59$ & $10$ & Right-arm joint representation \\
        $60$--$69$ & $10$ & Right TCP / end-effector representation \\
        $70$ & $1$ & Right gripper \\
        $71$--$99$ & $29$ & Right dexterous hand \\
        $100$--$127$ & $28$ & Reserved; always inactive \\
        \bottomrule
    \end{tabular*}
\end{table}

Unused coordinates, including unsupported body parts and reserved slots,
are zero-filled and marked invalid. For example, an arm with seven supplied
joint coordinates occupies seven entries of its ten-dimensional joint group;
the remaining three entries are inactive. After normalization, inactive
coordinates are reset to zero, and the data pipeline supplies explicit
state and action masks alongside the corresponding tensors. The canonical
layout therefore preserves the distinction between an unavailable quantity
and a valid coordinate whose value happens to be zero.

\subsubsection{Tactile Pretraining}
\label{app:tactile-pretraining}

\noindent\textbf{Data mixture.}
Both tactile pretraining stages use OpenNeoData~\citep{team2026n_0vtla},
the T-REX dataset~\citep{niu2026t}, and FTP-1-Sharpa~\citep{yuan2026ftp},
with sampling weights of $30\%$, $40\%$, and $30\%$, respectively.
The first stage uses this tactile-data mixture exclusively. In the second
stage, tactile data account for $90\%$ of the training mixture, retaining
the same relative weights across the three sources. The remaining $10\%$
consists of tactile-free data from world-action backbone pretraining and
supplements training of the visual and base-action pathways.
Visual inputs use a resolution of $192\times256$ pixels, and tactile
prediction targets span $24$ future frames at $15$ frames per second.
Table~\ref{tab:tactile-pretraining-config} summarizes the key settings.

\noindent\textbf{Stage 1: Tactile world model pretraining.}
We initialize the visual conditioning stream from the pretrained
world-action backbone and keep the LTX-Video model frozen. Only the
tactile world model is optimized, using a flow-matching prediction
objective in the shared VAE latent space. Visual features provide
conditioning through the unidirectional attention pathway, with gradients
stopped at the visual stream. The video prediction loss is disabled in
this stage. We sample tactile noise levels using a logit-normal sampler
with logit mean $0$ and standard deviation $1$, and train for $50{,}000$
optimizer steps with a global batch size of $1{,}024$.

\noindent\textbf{Stage 2: Tactile correction pretraining.}
We load the tactile world model weights learned in Stage 1 and initialize
the video world model, Q-Former, and action expert from the pretrained
world-action backbone described in Appendix~\ref{app:wa-pretraining}.
In particular, the action expert retains its pretrained action-generation
weights. We introduce the tactile expert with a zero-initialized
delta-velocity output head, and jointly train all four modules and the
Q-Former using the visual prediction, tactile prediction, base-action,
and steering objectives defined in the main method. The unidirectional
attention and gradient-isolation mechanisms remain in place during this
joint optimization.

The tactile world model supplies its final-block hidden states to the
tactile expert, which combines them with real-time tactile features to
learn local corrections to the base flow-matching velocity.
Tactile prediction training uses the noise endpoint $\sigma=1$ in this
stage, supporting single-step prediction. Actions are sampled at $30$ Hz
with a pretraining horizon of $50$ steps. Real-time tactile observations
are sampled at offsets of $0$, $4$, $8$, or $12$ action steps, and the
steering loss is evaluated over a $10$-step window following the sampled
offset. Relative-time encoding identifies the position of each action
with respect to the feedback time. We train for $50{,}000$ optimizer steps
with a global batch size of $512$.

Both stages use AdamW with $(\beta_1,\beta_2)=(0.9,0.95)$ and
$\epsilon=10^{-8}$, gradient clipping at $1.0$, and cosine learning-rate
decay with a minimum learning rate of $10^{-6}$.

\begin{table}[!htb]
    \centering
    \caption{\textbf{Key settings for the two tactile pretraining stages.}
    Stage 1 trains only the tactile world model; Stage 2 jointly trains
    all four modules and the Q-Former. Batch sizes are global; dashes
    indicate inapplicable settings.}
    \label{tab:tactile-pretraining-config}
    \small
    \setlength{\tabcolsep}{4pt}
    \renewcommand{\arraystretch}{1.12}
    \begin{tabular*}{\textwidth}{@{\extracolsep{\fill}}lcc@{}}
        \toprule
        Setting & Stage 1 & Stage 2 \\
        \midrule
        Training steps & $50{,}000$ & $50{,}000$ \\
        Global batch size & $1{,}024$ & $512$ \\
        Tactile / tactile-free sampling & $100\%$ / $0\%$ & $90\%$ / $10\%$ \\
        Video model and Q-Former learning rate & -- & $10^{-5}$ \\
        Tactile world model learning rate & $10^{-4}$ & $5\times10^{-5}$ \\
        Action expert learning rate & -- & $5\times10^{-5}$ \\
        Tactile expert learning rate & -- & $10^{-4}$ \\
        Warmup steps & $5{,}000$ & $1{,}000$ \\
        Weight decay & $0$ & $10^{-8}$ \\
        Tactile noise sampling & Logit-normal & Endpoint ($\sigma=1$) \\
        \bottomrule
    \end{tabular*}
\end{table}

\subsection{Additional Experimental Details}
\label{app:experimental-details}

\subsubsection{Real-Robot Platform and Teleoperation}
\label{app:real-robot-setup}

\begin{figure}[!htb]
    \centering
    \includegraphics[width=\textwidth]{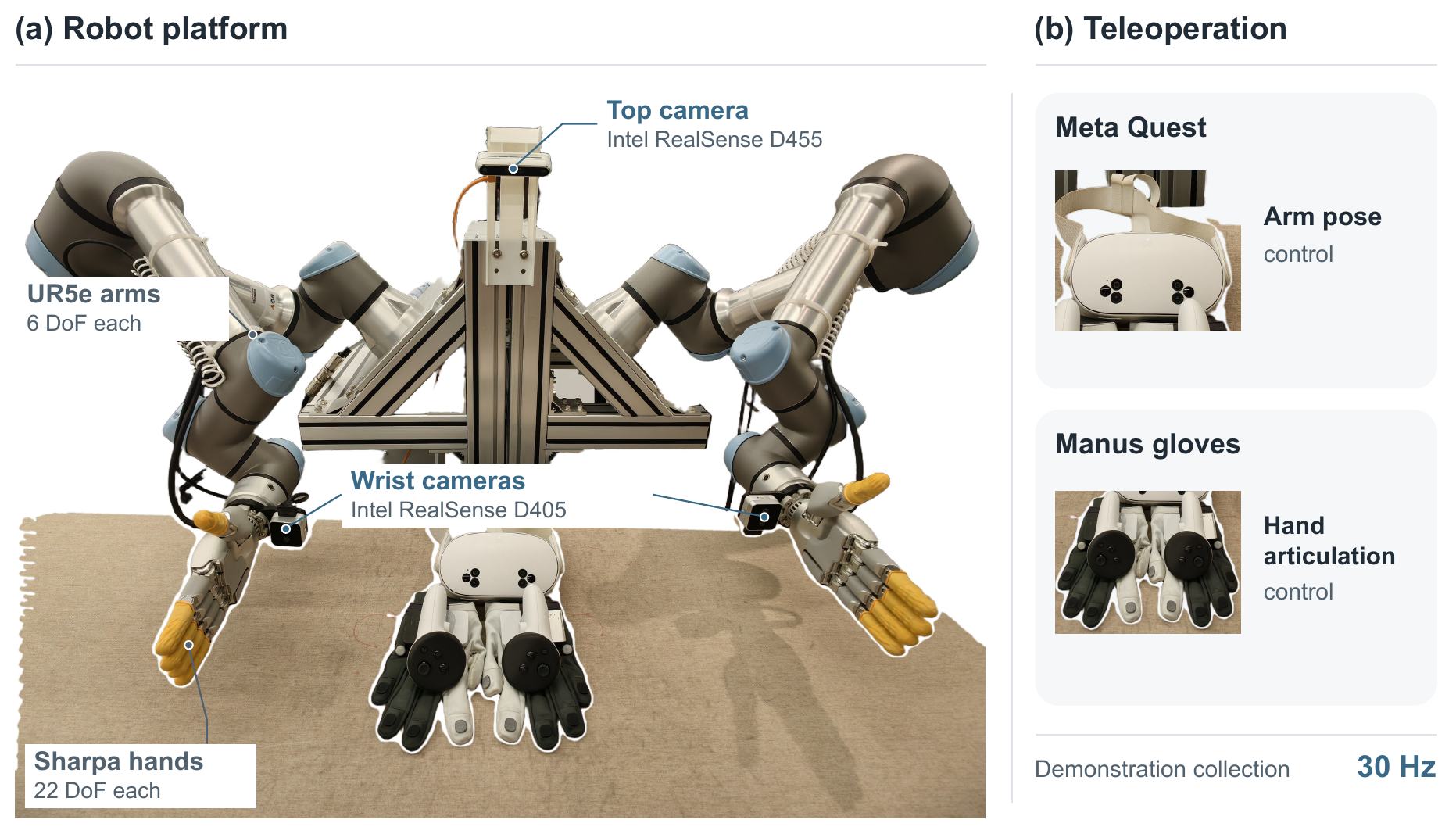}
    \caption{\textbf{Real-robot platform and demonstration collection.}
    \textbf{(a)} Two UR5e arms equipped with Sharpa dexterous hands share a
    tabletop workspace. A top-mounted Intel RealSense D455 and two
    wrist-mounted D405 cameras provide complementary visual observations.
    \textbf{(b)} Meta Quest supplies arm pose commands, while Manus gloves
    control hand articulation. Demonstrations are collected at $30$ Hz.}
    \label{fig:real-robot-setup}
\end{figure}

\noindent\textbf{Robot and sensing hardware.}
As shown in Fig.~\ref{fig:real-robot-setup}(a), our bimanual platform consists
of two 6-DoF UR5e robotic arms, each equipped with a 22-DoF Sharpa dexterous
hand. Both arms operate over a shared tabletop workspace. The visual
observation system comprises three cameras: one Intel RealSense D455
mounted above the workspace and one Intel RealSense D405 mounted at each
wrist. The top camera provides an overall view of the scene, while the
wrist cameras provide local views near the hands and manipulated objects.
These visual observations complement the tactile feedback acquired during
contact interactions.

\noindent\textbf{Teleoperation interface.}
We collect demonstrations using Manus gloves and a Meta Quest system
(Fig.~\ref{fig:real-robot-setup}(b)). The Manus gloves capture the operator's
finger articulation to control the Sharpa hands, while tracked motion from
the Meta Quest controllers provides pose commands for the UR5e arms.
This interface allows the operator to coordinate arm motion and finger
articulation during bimanual manipulation.

\noindent\textbf{Demonstration acquisition.}
Demonstrations are collected at $30$ Hz and contain multi-view visual
observations, tactile observations, robot states, and action commands.
We collect $100$ demonstrations for each of the six real-world tasks
described in the main text, yielding $600$ demonstrations in total.

\subsubsection{Simulation Training and Evaluation}
\label{app:simulation-evaluation}

We evaluate on UniVTAC~\citep{chen2026univtac}. For methods trained in our
experiments, we partition the six tasks according to their visual inputs
and train one checkpoint jointly on the tasks within each group.
The single-view group consists of \emph{Lift Bottle}, \emph{Pull Out Key},
\emph{Put Bottle}, and \emph{Insert Hole}; the multi-view group consists of
\emph{Lift Can} and \emph{Insert Tube}.
Each method therefore uses two group-level checkpoints, with each
checkpoint evaluated on all tasks in its corresponding group.

We follow the official UniVTAC evaluation protocol and task-success
criteria. Each method evaluated by us is tested for $50$ trials per task.
The task-level success rate is the fraction of successful trials,
expressed as a percentage. The overall score is the unweighted mean of
the six task-level success rates.
The ACT and VITaL scores in Table~\ref{tab:univtac-main} are reproduced
from UniVTAC. The two-checkpoint training setup and $50$-trial evaluation
described here apply to methods trained and evaluated by us, rather than
these published reference results.

\subsubsection{Real-World Training and Baseline Adaptation}
\label{app:real-world-training}

\noindent\textbf{Training data and baseline fine-tuning.}
Each real-world task provides $100$ demonstrations collected at $30$ Hz
using the platform described in Appendix~\ref{app:real-robot-setup}.
For T-Rex~\citep{niu2026t} and FTP-1~\citep{yuan2026ftp}, we follow their
respective official training procedures and perform $20{,}000$
task-specific optimization steps on the collected demonstrations.

\noindent\textbf{Adapting the $\pi_{0.5}$ action interface.}
The released $\pi_{0.5}$ checkpoint used in our experiments has a
32-dimensional action interface, which is insufficient to accommodate
the arm and dexterous-hand commands of our UR5e--Sharpa platform.
We expand its linear action projection to support an 80-dimensional
action representation. TCP commands retain their assigned positions
within the original first $32$ dimensions, while the Sharpa hand-joint
commands are assigned to the newly added dimensions. With zero-based
indexing, the original action subspace occupies indices $0$--$31$ and
the added hand-command subspace occupies indices $32$--$79$.
This 80-dimensional interface is specific to the $\pi_{0.5}$ adaptation;
the canonical layout used by our backbone is described in
Table~\ref{tab:canonical-128}.

\subsubsection{Real-World Evaluation and Progress Scoring}
\label{app:real-world-evaluation}

We evaluate each method for $20$ trials on each real-world task.
Each trial receives a normalized progress score $p_i\in[0,1]$ according
to the task-specific milestones in Table~\ref{tab:progress-scoring}.
For sequential tasks, the listed values are cumulative scores: a trial
receives the score of the furthest completed milestone, and scores are
not summed across milestones. A trial that reaches none of the listed
milestones receives $0$. For fruit placement, the score increases by
$0.25$ for each of the four objects successfully placed on the plate.
The progress success rate (PSR) for a task is
\begin{equation}
    \operatorname{PSR}
    = \frac{100}{N}\sum_{i=1}^{N}p_i,
    \qquad N=20.
\end{equation}
We report PSR as a percentage, and compute the overall real-world score
as the unweighted mean of the six task-level PSRs.

\begin{table}[!htb]
    \centering
    \caption{\textbf{Progress scoring for real-world evaluation.}
    Scores range from $0$ to $1$ for each trial. Sequential-task scores
    indicate cumulative progress, rather than additive rewards.}
    \label{tab:progress-scoring}
    \small
    \setlength{\tabcolsep}{5pt}
    \renewcommand{\arraystretch}{1.2}
    \begin{tabular}{@{}>{\raggedright\arraybackslash}p{0.21\textwidth}
                       >{\raggedright\arraybackslash}p{0.75\textwidth}@{}}
        \toprule
        Task & Progress milestones and scores \\
        \midrule
        Fruit on Plate
        & $0.25$ for each object successfully placed on the plate;
          $1.0$ when all four objects are placed. \\
        \addlinespace[3pt]
        Close Zipper
        & $0.2$: the left hand lifts the bag;
          $0.6$: the right hand begins pulling the zipper, which is not
          yet fully closed;
          $1.0$: the zipper is fully closed. \\
        \addlinespace[3pt]
        Wipe Whiteboard
        & $0.2$: the robot grasps the eraser;
          $0.6$: it performs an initial wiping stroke on the board;
          $1.0$: the board is wiped clean. \\
        \addlinespace[3pt]
        Twist Cap
        & $0.2$: the left hand lifts the bottle;
          $0.6$: the robot successfully rotates the cap;
          $1.0$: the cap is fully unscrewed. \\
        \addlinespace[3pt]
        Split Cups
        & $0.2$: the robot lifts the cups;
          $0.6$: it inserts its fingers into a cup;
          $1.0$: one cup is separated from the others. \\
        \addlinespace[3pt]
        Dropper Transfer
        & $0.2$: the robot grasps the dropper;
          $0.8$: liquid is transferred into the beaker;
          $1.0$: the dropper is put back after the transfer. \\
        \bottomrule
    \end{tabular}
\end{table}

\subsection{Ablation Details}
\label{app:ablation-details}

\subsubsection{Model Design Ablations}
\label{app:model-design-ablation}

\noindent\textbf{Initialization and post-training.}
All model-design variants in Table~\ref{tab:ablation} are initialized
from the same checkpoint obtained after Stage 2 tactile correction
pretraining (Appendix~\ref{app:tactile-pretraining}). We apply the
corresponding architectural or objective changes before task-specific
post-training and train each variant separately. The reported results
therefore evaluate models adapted to their ablated configurations,
rather than inference-time modifications of a fully post-trained model.
We use the progress scoring protocol in
Appendix~\ref{app:real-world-evaluation}.

\noindent\textbf{Without both tactile modalities.}
We remove all tactile-specific modules and computation paths, including
the tactile world model, real-time tactile encoding, and tactile expert.
The retained video world model, Q-Former, and action expert undergo
world-action backbone post-training. Actions are generated entirely
through the base action expert, without tactile prediction, feature
fusion, or delta-velocity refinement.

\noindent\textbf{Without predicted or real-time tactile.}
For \emph{w/o predicted tactile}, we disable the predicted-tactile
subexpert and retain the real-time tactile subexpert. For
\emph{w/o real-time tactile}, we disable the real-time tactile subexpert
and retain the predicted-tactile subexpert. In both cases, we bypass
sigmoid weighting and pass the active subexpert's feature directly to
the subsequent tactile-expert computation. The remaining branch thus
continues to support delta-velocity prediction without a weighted
combination of the two tactile sources.

\noindent\textbf{Action-space residual correction.}
We retain the same model architecture but change the correction target
and the point at which the correction is applied. The action expert
first completes flow-matching generation to produce the base action
sequence. The tactile branch then predicts an action-space residual
that corrects this final sequence. During post-training, we replace
the velocity-steering loss with a residual loss on the final actions.
This variant compares refinement during flow-matching generation with
correction after base-action generation using the same architecture.

\noindent\textbf{Without asynchronous inference.}
We set the training delay to $d=0$ and remove the local correction-window
restriction, evaluating the steering loss over all $H$ actions in the
chunk. At inference, the world-model planning system and the
action-generation system operate at the same update frequency:
world priors and action chunks are refreshed synchronously. This
variant removes the separation between low-frequency planning and
high-frequency tactile refinement during both post-training and
execution.

\noindent\textbf{Without attention masking and stop-gradient.}
We jointly remove the attention-mask and stop-gradient constraints
between the action expert and tactile expert, and independently
post-train this variant from the same Stage 2 checkpoint. Both the
forward attention restriction and the gradient-isolation mechanism
are removed during post-training. This experiment evaluates their
combined role in separating base action generation from tactile
refinement.

\subsubsection{Pretraining Ablation}
\label{app:pretraining-ablation}

The \emph{w/o pretrain} variant in Fig.~\ref{fig:pretraining-ablation}(a)
uses the original world-action checkpoint obtained before either tactile
pretraining stage (Appendix~\ref{app:wa-pretraining}). The video world
model, Q-Former, and action expert retain these pretrained weights,
while the tactile world model and tactile expert are randomly
initialized. We skip both tactile world model pretraining and tactile
correction pretraining, and proceed directly to task-specific
post-training using the full model's joint objective.

Both tactile modules and both tactile sources remain available in this
variant during post-training and inference. The comparison therefore
evaluates the contribution of the two tactile pretraining stages on top
of the original world-action initialization. The fully pretrained
counterpart starts task-specific post-training from the checkpoint
obtained after Stage 2.

\subsection{Real-World Policy Rollouts}
\label{app:policy-rollouts}

\begin{figure}[!htb]
    \centering
    \includegraphics[width=\textwidth]{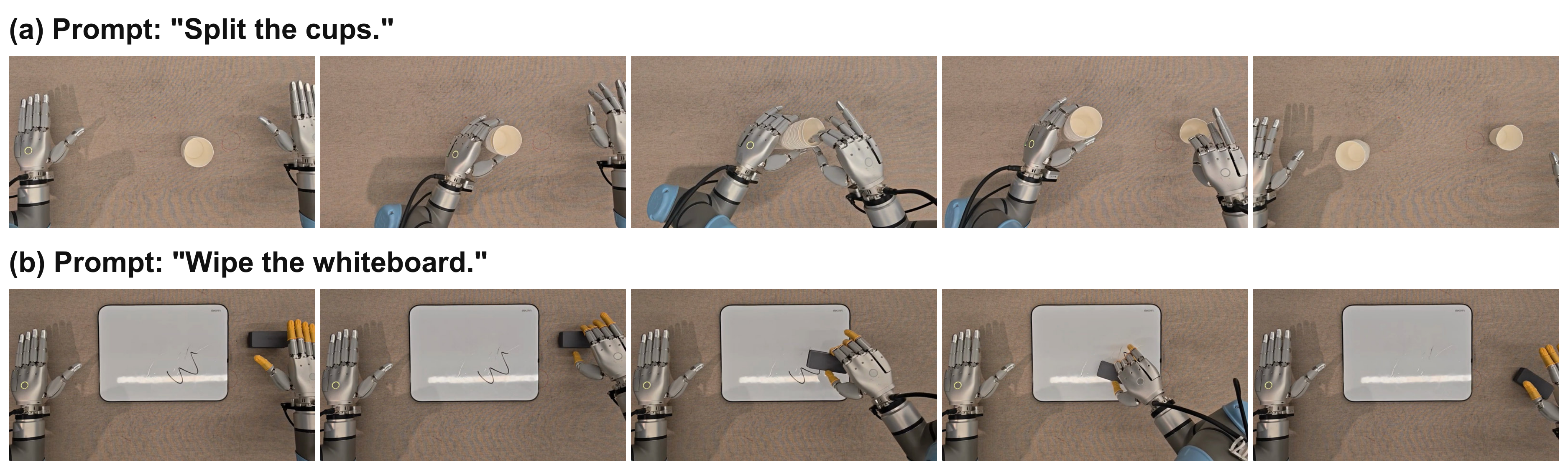}
    \caption{\textbf{Representative real-world policy rollouts.}
    \textbf{(a)} Split Cups. \textbf{(b)} Wipe Whiteboard.
    Each row shows five selected frames in chronological order from left
    to right, with the task prompt at the upper left.}
    \label{fig:policy-rollouts}
\end{figure}

Figure~\ref{fig:policy-rollouts} illustrates the progression of two
contact-rich tasks during policy execution. In \emph{Split Cups}, the
robot grasps the nested cups, engages them with the other hand, and
separates a cup before placing the cups apart. In \emph{Wipe Whiteboard},
the robot grasps the eraser, brings it into contact with the board,
wipes across the marked region, and moves the eraser away after wiping.
These selected frames complement the quantitative evaluations in the
main text. Full executions are available in the supplementary videos.